\pdfoutput=1

\documentclass[11pt]{article}

\usepackage[]{acl}

\usepackage{times}
\usepackage{latexsym}

\usepackage[T1]{fontenc}

\usepackage[utf8]{inputenc}

\usepackage{microtype}

\usepackage{inconsolata}

\usepackage{multirow, booktabs}
\usepackage{makecell, hhline}
\usepackage{xspace}
\usepackage{subfigure}
\usepackage{pifont}
\usepackage{longtable}
\usepackage[para,online,flushleft]{threeparttable}
\usepackage{algorithm}
\usepackage{algpseudocode}
\usepackage{amsmath,amssymb,mathtools,amsfonts,amsthm}
\usepackage{siunitx} 
\definecolor{hookersgreen}{rgb}{0.0, 0.44, 0.0}
\definecolor{indiagreen}{rgb}{0.07, 0.53, 0.03}
\definecolor{islamicgreen}{rgb}{0.0, 0.56, 0.0}
\definecolor{kellygreen}{rgb}{0.3, 0.73, 0.09}
\definecolor{alizarin}{rgb}{0.82, 0.1, 0.26}
\newcommand{\cmark}{{\color{kellygreen} \ding{51}}}
\newcommand{\xmark}{{\color{alizarin} \ding{55}}}

\algnewcommand\algorithmicinput{\textbf{Input:}}
\algnewcommand\algorithmicoutput{\textbf{Output:}}
\algnewcommand\algorithmicparameter{\textbf{Parameters:}}
\algnewcommand\INPUT{\item[\algorithmicinput]}
\algnewcommand\OUTPUT{\item[\algorithmicoutput]}
\algnewcommand\PARAMETER{\item[\algorithmicparameter]}

\DeclareMathOperator*{\argmax}{arg\,max}

\DeclareMathOperator{\Loss}{\mathcal{L}}
\DeclareMathOperator{\Dataset}{\mathcal{D}}
\DeclareMathOperator{\Corpus}{\mathcal{C}}

\DeclareMathOperator{\Prob}{\mathcal{P}}
\DeclareMathOperator{\Candidates}{\mathcal{Z}}
\DeclareMathOperator{\Encoder}{\mathbf{E}}
\DeclareMathOperator{\logits}{\mathbf{q}}

\newcommand{\method}{RCR\xspace}

\title{Reinforcing Compositional Retrieval: Retrieving Step-by-Step for \\Composing Informative Contexts}

\author{Quanyu Long\textsuperscript{\rm *1}~~ Jianda Chen\textsuperscript{\rm *1} ~~ Zhengyuan Liu\textsuperscript{\rm 2}~~ Nancy F. Chen\textsuperscript{\rm 2}~~ \\
{\bf  Wenya Wang\textsuperscript{\rm 1}~~ Sinno Jialin Pan\textsuperscript{\rm 1,3} } \\
\textsuperscript{\rm 1}Nanyang Technological University, Singapore\\
\textsuperscript{\rm 2}Institute for Infocomm Research (I2R), A*STAR, Singapore \\
\textsuperscript{\rm 3}The Chinese University of Hong Kong~~ \\
\texttt{\{quanyu001, jianda001\}@ntu.edu.sg}}

\begin{document}
\maketitle

\def\thefootnote{*}\footnotetext{Equal contribution.}
\def\thefootnote{\arabic{footnote}}

\begin{abstract}
Large Language Models (LLMs) have demonstrated remarkable capabilities across numerous tasks, yet they often rely on external context to handle complex tasks. While retrieval-augmented frameworks traditionally focus on selecting top-ranked documents in a single pass, many real-world scenarios demand compositional retrieval, where multiple sources must be combined in a coordinated manner. In this work, we propose a tri-encoder sequential retriever that models this process as a Markov Decision Process (MDP), decomposing the probability of retrieving a set of elements into a sequence of conditional probabilities and allowing each retrieval step to be conditioned on previously selected examples. We train the retriever in two stages: first, we efficiently construct supervised sequential data for initial policy training; we then refine the policy to align with the LLM’s preferences using a reward grounded in the structural correspondence of generated programs. 
Experimental results show that our method consistently and significantly outperforms baselines, underscoring the importance of explicitly modeling inter-example dependencies. These findings highlight the potential of compositional retrieval for tasks requiring multiple pieces of evidence or examples\footnote{Codes of this work are available at \url{https://github.com/ruyue0001/Step-by-Step-Retrieval}}.
\end{abstract}

\section{Introduction}
\label{sec:intro}
Large Language Models (LLMs) have demonstrated remarkable progress in recent years, mastering text generation and diverse problem-solving. Context often plays a critical role in grounding these models in task-relevant information: it helps incorporate domain-specific knowledge, clarify ambiguous queries, and provide demonstrative examples. This leads to the common adoption of retrieval-augmented frameworks \citep{lewis2020retrieval,izacard2022few,liu2022makes}, which introduce a ``retrieve'' step to extract relevant context segments such as documents or examples to enrich the LLM's input, improving both the accuracy and reliability of its outputs.


\begin{figure}[t]
    \centering
    \includegraphics[width=0.95\columnwidth]{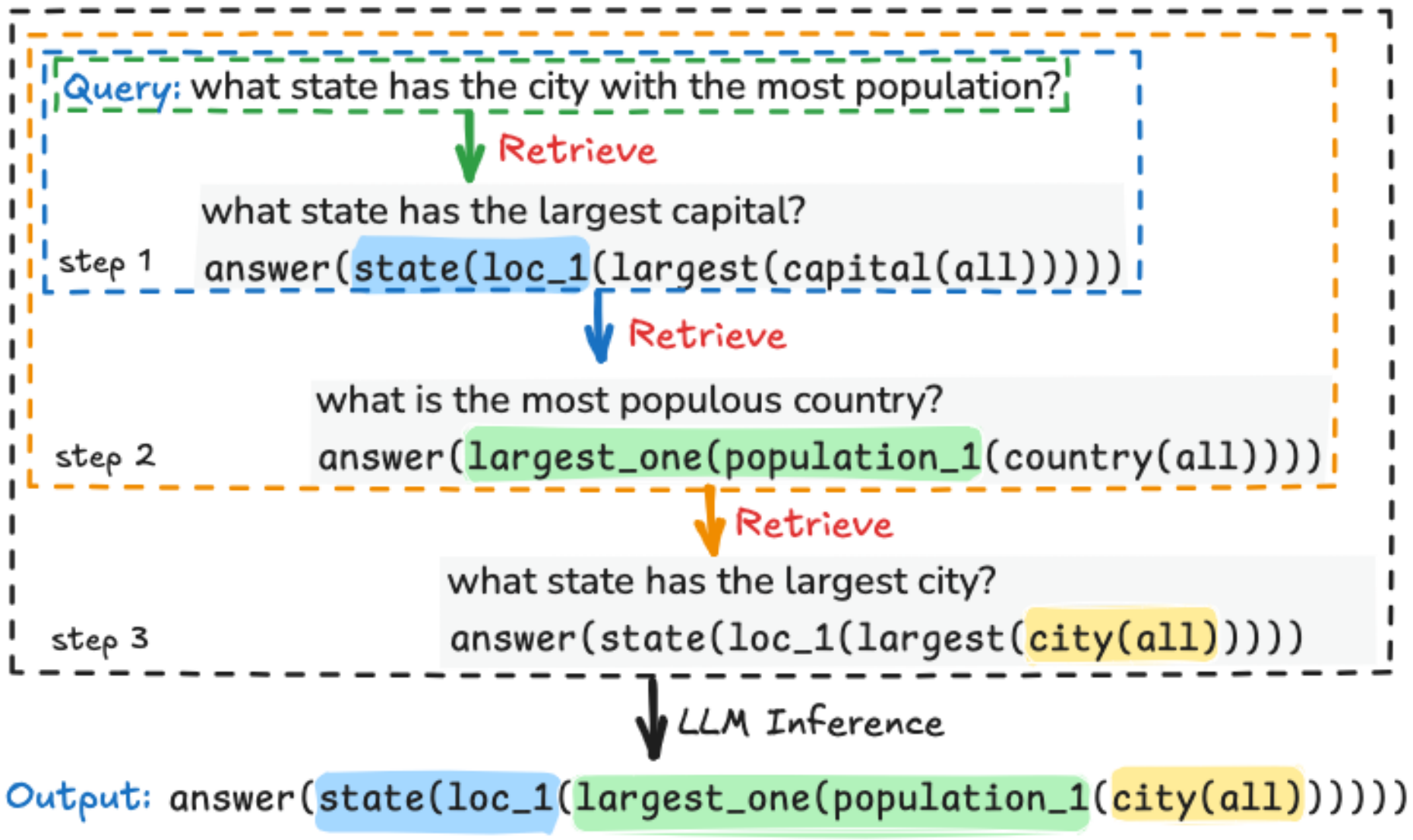}
    \caption{When generating the target semantic program of a query (parsing the query into the logical form), relying on single or repetitive demonstrative context cannot cover the ground-truth program local structure and lead to poor context augmentation quality. In order to capture the whole retrieval results’ collective effect to facilitate better local structure coverage and program generalization, we solve this compositional retrieval problem by modeling it as a Markov Decision Process.
    }
    \label{fig:moti}
\end{figure}

However, many complex tasks demand combining multiple pieces of evidence or examples of diverse semantics or structures. For instance, multi-hop queries often require several documents to accumulate all necessary insights. Another scenario, as illustrated in Figure \ref{fig:moti}, involves translating a query into a logical forms, where an LLM needs to assemble symbols or sub-trees from multiple demonstrative examples to construct the target program accurately. In this case, relying on a single retrieved example can only provide partial hints, while repetitive or redundant examples provide little additional information. Consequently, this compositional retrieval setting necessitates coordinated selection of examples to maximize collective utility of chosen composition.

To address these complexities, conventional top-$k$ retrieval is insufficient, as it operates in a single-turn manner and fails to capture how newly selected items condition subsequent choices. Therefore, we propose a compositional retrieval paradigm that conducts retrieval in multiple steps, with each step adapting to the evolving context formed by previously retrieved items. This approach ensures that newly selected examples complement previously chosen ones, improving the overall coverage of retrieved content (Figure \ref{fig:moti}).
Although recent works have explored sequential retrieval for selecting exemplary contexts \citep{liu2024se2sequentialexampleselection,liu2024demorank}, they typically rely on a large scoring language model for constructing training sequences, incurring significant computational overhead. 

To maximize the collective utility of the retrieved examples instead of selecting multiple high-scoring yet potentially redundant elements from a single global ranking, we formulate compositional retrieval as a Markov Decision Process (MDP), wherein each state reflects both the query and the previously chosen items. We adopt a tri-encoder architecture as the policy model, mapping each query and partial retrieval sequence to a probability distribution over candidate examples.
To initialize this policy, we construct supervised fine-tuning data in a more efficient manner by maximizing sub-structure coverage without scoring models. In the second stage, we perform reinforcement learning (RL) and adopt a grouped policy optimization \citep{shao2024deepseekmath,deepseek2025deepseek}, reinforcing the policy using a task-specific reward based on local structure coverage of the generated logical form. This design captures the LLM’s compositional needs and progressively enhances the quality of the assembled context.

We evaluate our method \textbf{R}einforcing \textbf{C}ompositional \textbf{R}etrieval (\textbf{RCR}) on compositional generalization semantic parsing benchmarks which requires generating new combinations of familiar structures and symbols, making it an ideal initial scenario to examine how context compositions impact program generation capabilities. Results show our method consistently outperforms top-$k$ and sequential retrieval baselines.
The observed improvements stem from our tri-encoder retriever’s ability to model inter-example dependencies, enabling more effective context composition and enhancing program generation quality within the LLM.
Our analysis further underscores the effectiveness of leveraging RL in refining retrieval strategies. We systematically evaluate different RL setups and advantage estimation methods, providing a comprehensive evaluation and insights into both its benefits and limitations in retrieval tasks.
In summary, our key contributions are:
\begin{itemize}
    \item We formalize a compositional retrieval process that explicitly accounts for coordinated selection of contexts via sequential conditional probabilities.
    \item We develop a tri-encoder retriever, modeling retrieval as a Markov Decision Process and factoring each selection step upon previously chosen examples.
    \item We introduce an efficient supervised fine-tuning step for data construction and subsequently refine the retriever through reinforcement learning, aligning it with the LLM’s downstream performance.
\end{itemize}

\section{Problem Formulation}
\label{sec:compo}
In this work, we depart from conventional top-ranked retrieval which treat each candidate independently. Instead, we propose a compositional retrieval problem that aims to optimize the selection of the most influential context group, and each element within this group will contribute together to lead to a correct prediction.

\subsection{Compositional Retrieval}
The common retrieval-augmented framework mainly contains two components, a dense retriever to retrieve relevant contexts and a Large Language Model (LLM) to accept retrieved contexts for generating responses. Given a set of retrieved elements $\Candidates$, LLMs will condition on the combination of context $\Candidates$ and query $x$ to generate the output $y$ by prepending all $k$ retrieved elements to $x$:
\begin{equation}
    \Prob(y\mid x)\approx\sum\limits_{\Candidates} \Prob_{\rm LM}(y\mid \Candidates,x)\Prob(\Candidates\mid x),
    \label{equation:rag}
\end{equation}
where $\Candidates=[z_{1},\cdots,z_{k}]$, and $[\cdot]$ denotes text concatenation. In order to retrieve $k$ elements, in standard dense retrieval, the likelihood of retrieving a single candidate $z$ is computed as:
\begin{equation}
    \Prob(z\mid x)\approx\frac{\exp(\text{sim}(x,z))}{\sum\limits_{z^{\prime}\in\Corpus}\exp(\text{sim}(x,z^{\prime}))},
    \label{equation:dense_single}
\end{equation}
where $\text{sim}(\cdot,\cdot)$ is a similarity function, which is often computed via inner product, $\text{sim}(x,z)=\Encoder_{c}(z)^{\top}\Encoder_{q}(x)$. The final context $\Candidates$ is obtained by selecting the top-$k$ candidates based by Eq.~\eqref{equation:dense_single}.

However, such top-$k$ selection treats candidates independently,\footnote{The selection of each element from a probability distribution is not necessarily independent when considering rank dependence, i.e. once choosing one element, the remaining elements are selected from a constrained set without replacement using re-normalization. However, once considering rank dependence, it inherently become a sequential selection.} disregarding the compositional quality of the retrieved set and failing to assess the joint utility of retrieved elements, which is crucial for tasks where multiple documents must interact to form an informative context.
To address this, we define compositional retrieval as computing the probability of retrieving a set of candidates as a sequence of conditional probabilities:
\begin{equation}
\Prob(\Candidates| x) = \Prob(z_{1}| x) \prod_{i=2}^k \Prob(z_{i} | x, z_{1}, \dots, z_{i-1}).
\label{equation:sequential}
\end{equation}

We explicitly model dependencies among retrieved elements, where each selection influences subsequent choices. Unlike traditional methods that independently select top ranked elements in a single turn and unable to estimate the compositional retrieval probability, this approach optimizes the overall composition by accounting for both individual relevance and inter-example interactions, leading to more collective context selection.

\subsection{Overview of the MDP Modeling}

\begin{figure}[t]
    \centering
    \includegraphics[width=0.95\columnwidth]{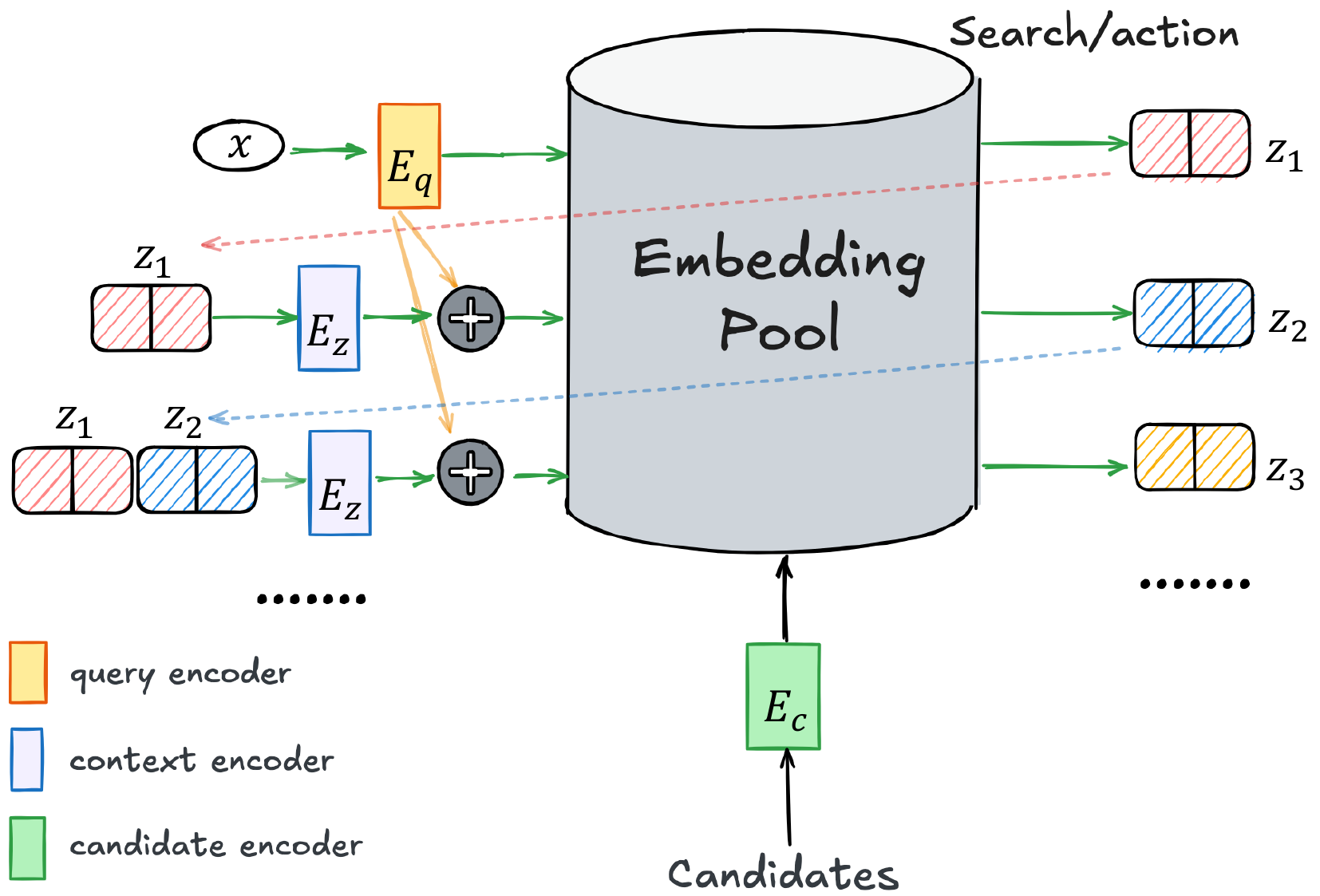}
    \caption{Tri-encoder model to retrieve step-by-step.}
    \label{fig:seq_retr}
\end{figure}

We further formalize the proposed step-by-step retrieval process as a Markov Decision Process (MDP). In compositional program generalization (an example in Figure \ref{fig:moti}), prior works \citep{levy-etal-2023-diverse,an2023context} show that increasing structural coverage (i.e., sub-trees of the program tree) of the expected program within the demonstrative context improves in-context learning (ICL) performance, since high-coverage contexts enhance generalization by exposing the model to essential program symbols (e.g., predicates and logical operators). 
Thus, the goal of our decision process is to select program examples that provide more expected symbols and larger structures coverage within the whole context, while minimizing redundant selections with limited utility.

\paragraph{State} At each step $t$, state is defined as $s_{t}=[x,z_{1},\cdots,z_{t-1}]$, representing the partial retrieval sequence, where $x$ is the input, and $z_{i}$ is $i$-th selected program example.

\paragraph{Policy} To compute the conditional probability in Eq.~\eqref{equation:sequential}, we avoid using a single encoder to process the concatenated input $[x,z_{1},\cdots,z_{t-1}]$, which may weaken the signal input $x$.
Instead, we employ a query encoder $\Encoder_{x}$ and a context encoder $\Encoder_{z}$ to separately encode $x$ and each selected example $z_{i}$ (Figure \ref{fig:seq_retr}).
Given a candidate pool $\Corpus=\{c_{j}\}^{N}_{j=1}$, we introduce a third encoder $\Encoder_{c}$ to encode candidate examples, the selection logit for candidate $c_{j}$ at step $t$ are computed as:
\begin{equation}
    {\rm q}(x,\Candidates_{t-1},c_{j}) = \Encoder_{c}(c_j)^{\top}(\Encoder_{x}(x)+\lambda \sum_{i=1}^{t-1}\Encoder_{z}(z_{i})),
    \label{equation:logits}
\end{equation}
where $\lambda$ is a weighting factor, and $\Candidates_{t-1} = [z_{1},\cdots,z_{t-1}]$ represents selected list at time step $t-1$. The policy distribution is then given by: $\pi_{\theta}(\cdot|x, z_{i<t}) = {\rm softmax} (\logits(x,\Candidates_{t-1},\cdot)/\tau)$, $\tau$ is a scaling temperature.

\paragraph{Action} At step $t$, an action involves selecting an example $z_{t}$ from $\Corpus$ by sampling from $\pi_{\theta}(\cdot|x, z_{i<t})$. Once selected, an action is removed from the candidate pool to prevent duplicate selection.

\paragraph{Reward}
After $k$ retrieval steps, the selected examples serve as context for prompting an LLM in an ICL manner to generate a response $\hat{y}$:
\begin{equation}
    \hat{y}_{i}\sim\Prob_{\rm LLM}(\hat{y}_{i}|[\underbrace{z_{1},z_{2},\cdots,z_{k}}_{\rm context},x],\hat{y}_{1:i-1}).
    \label{equation:inference}
\end{equation}

We define a reward based on local structural similarity between the generated and reference programs. Following \citet{levy-etal-2023-diverse}, we first anonymize programs by replacing values (e.g., strings, numbers) with constants, as these are typically irrelevant for structural coverage. Each program is then represented as a tree, where nodes correspond to program symbols and edges capture parent-child relationships between functions and their arguments.
A local structure is defined as a subgraph where all leaf nodes share a sibling relationship, meaning it includes only parent-child and sibling connections while excluding cousin or uncle relations. The size of a local structure $l$ is defined as the number of nodes in this sub-graph (see Figure \ref{fig:local_structure}). Additional details are provided in Appendix \ref{sec:appendixB}, with examples in Table \ref{tab:ls_examples}.

\begin{figure}[t]
    \centering
    \includegraphics[width=0.95\columnwidth]{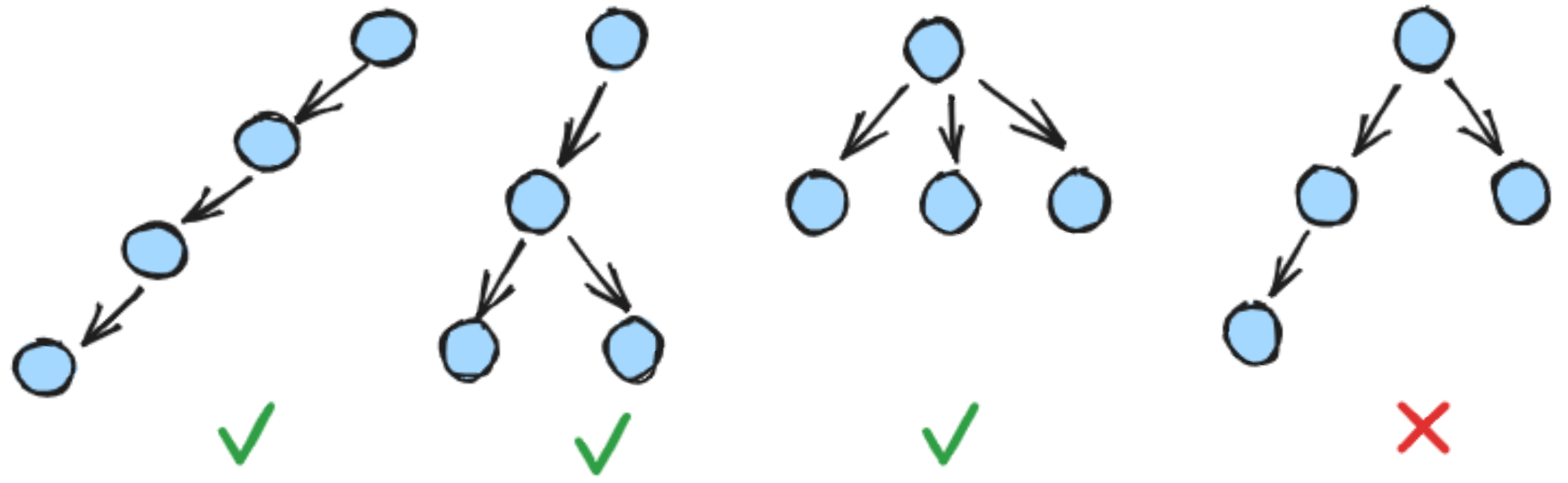}
    \caption{Local structures when size $l=4$. Following \citet{levy-etal-2023-diverse}, we exclude more distant familial ties.}
    \label{fig:local_structure}
\end{figure}

We extract local structures of size $\leq l$ from both the generated program $\hat{y}$ and the ground-truth program $y$, and then collect all the extracted structures into a set. We denote such extraction as a function $LS^{l}(\cdot)$. The reward is then computed as the Jaccard similarity between their local structures:
\begin{equation}
    r(o)= \frac{|LS^{l}(\hat{y})\cap LS^{l}(y)|}{|LS^{l}(\hat{y})\cup LS^{l}(y)|},
    \label{equation:reward}
\end{equation}
where $o=[x,z_{1},\cdots,z_{k}]$ is the final observation. This formulation naturally penalizes the generation of overly lengthy programs. 

\section{Reinforcing Compositional Retrieval}
\label{sec:rl}
The training of our propose tri-encoder sequential retriever contains two stages, supervised fine-tuning (SFT) and reinforcement learning (RL).

\begin{figure*}[ht]
\centering
\includegraphics[width=0.99\textwidth]{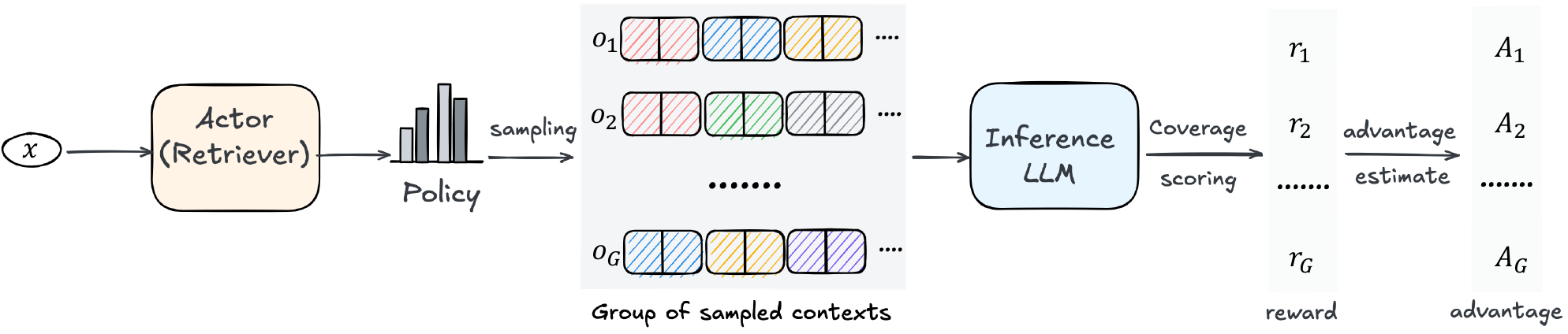}
\caption{Group Relative Policy Optimization (GRPO) process for optimizing the sequential retriever.}
\label{fig:grpo}
\end{figure*}

\subsection{Stage 1: SFT}

To construct sequential training data and establish a strong initial policy, we propose an intuitive and efficient data construction method. 
Given an input $x$ and a candidate pool $\Corpus$, we first construct the training data for step 1.
A high-quality program example should maximize local structure coverage, as broader coverage reduces the model’s need to generate new structures and provides insights into structural fusion.
Therefore, we greedily sample the candidate in the pool $\Corpus$ that maximizes local structure coverage relative to the ground-truth $y$.

For subsequent steps, intuitively, an optimal composition should introduce diverse symbols and local structures to enhance generalization. Therefore, retrieval should prioritize covering unseen structures. 
More specifically, as shown in Figure \ref{fig:moti}, after expanding the set of retrieved examples (encode by $\Encoder_{z}$ in Figure \ref{fig:seq_retr}), the retriever \textbf{should select examples that can cover previously uncovered structures} in $y$. We iteratively sample candidates from $\Corpus$ that maximize coverage of these remaining structures. Algorithm \ref{alg1} presents the data construction procedure for SFT stage. 

\begin{algorithm} [t]
    \caption{Data construction for SFT}
    \label{alg1}
    \begin{algorithmic}[1]
        \INPUT data instance $(x,y)$, example pool $\Corpus$, total step $k$, maximum local structure size $l$, local structure extractor $LS^{l}(\cdot)$, uncovered local structure set $\mathcal{S}$, selected examples list $\Candidates_{t}$ at step $t$.
        \State $\mathcal{S}\gets LS^{l}(y)$
        \State $\Candidates_{0}\gets []$
        \For {$t=1,\cdots,k$}
            \State $c_{t} \gets \argmax_{c}|LS^{l}(c)\cap \mathcal{S}|,c\in \Corpus$ 
            \State Append $(x,\Candidates_{t-1},c_{t})$ to $\Dataset^{SFT}$
            \State $z_{t}\gets c_{t}$
            \State $\Candidates_{t}\gets\Candidates_{t-1}+[z_{t}]$
            \State $\mathcal{S} \gets S-LS^{l}(z_{t})$
	\EndFor
        \OUTPUT Training Data $\Dataset^{SFT}$ for $(x,y)$.
    \end{algorithmic} 
\end{algorithm}

Unlike existing data construction methods that rely on a scoring LLM (often distinct from the inference model) to encode selected examples and rank training data based on similarity scores~\citep{liu2024se2sequentialexampleselection,liu2024demorank}, our approach eliminates the computational overhead of LLM forward passes while maintaining a more interpretable selection process.

After constructing the training data, each instance $(x,\Candidates_{t-1},c_{t})\in \Dataset^{SFT}$ is used to train the tri-encoder retriever. 
The objective is to maximize the selection logit of the candidate $c_{t}$, treating it as the positive example given input $x$ and previously retrieved examples $\Candidates_{t-1}$.
For negative samples $\mathcal{N}$, we incorporate both in-batch negatives and hard negatives. Hard negatives are selected using a two-step process: 1) retrieve top-$H$ most similar example to $x$ using BM25 \citep{robertson2009probabilistic}, 2) rank these $H$ candidates by their local structure coverage relative to the ground-truth $y$ and select the bottom-$B$ (high-similarity but low-coverage) candidates as hard negatives.
The tri-encoder is trained using the InfoNCE loss \citep{oord2018representation,chen2020simple}:
\begin{align}
    &\Loss(x,\Candidates_{t-1},c_{t},\mathcal{N})= \nonumber \\
    &-\log\frac{\mathrm{exp}\left({\rm q}(x,\Candidates_{t-1},c_{t})\right)}{ \sum_{c_{j} \in \{\mathcal{N} \bigcup c_{t} \}}\mathrm{exp}\left({\rm q}(x,\Candidates_{t-1},c_{j})\right)},
\end{align}
Where ${\rm q}(x,\Candidates_{t-1},c_{j})$ is computed by Eq.~\eqref{equation:logits}.

\subsection{Stage 2: RL}

Although the learned initial policy ensures high coverage of the expected output program, it does not guarantee correct generation by the inference LLM. The retriever should also capture the LLM’s preferences and align accordingly.
With a sequential retriever, we collect a chain of retrieved examples over $k$ steps and then feed the concatenation of the prompt and test utterance $[\Candidates_{k},x]$ to a LLM (Eq.~\eqref{equation:inference}). After generating the full response (Eq.~\eqref{equation:reward}), we obtain a reward. This reward signal eliminates the need for pairwise preference data, allowing reinforcement learning to optimize retrieval based on aggregated evaluations across multiple example chains. 

To achieve this, we adopt Group Relative Policy Optimization (GRPO) \citep{shao2024deepseekmath} , which enables efficient policy alignment with the LLM without requiring a separately trained value model. 
For each query $x$, we sample a group of outputs $\{o_{1},\cdots,o_{G}\}$ using the tri-encoder with the old policy $\pi_{\theta_{old}}$ and obtain a corresponding group of rewards $\{r_{1},\cdots,r_{G}\}$. The policy is then optimized by maximizing the following objective:
\begin{align}
    \mathcal{J}&(\theta)=\frac{1}{G}\sum_{i=1}^{G}\bigg( \min\Big(r_{\theta}(o_{i})A_{i},{\rm clip}\big(r_{\theta}(o_{i}), \nonumber \\
    &1-\epsilon,1+\epsilon\big)A_{i}\Big) - \beta D_{KL}(\pi_{\theta}\parallel \pi_{ref}) \bigg),
    \label{equation:grpo}
\end{align}
where $r_{\theta}(o_{i})$ is the probability ratio between the new and old policy, $r_{\theta}(o_{i})=\frac{\pi_{\theta}(o_{i}|x)}{\pi_{\theta_{old}}(o_{i}|x)}$. The clipping function ${\rm clip}$ truncates the ratio between the range $[1-\epsilon, 1+\epsilon]$, stabilizing updates. The advantage function $A_{i}$ computed using normalized group rewards:
\begin{equation}
   A_{i}=\frac{r_{i}-{\rm mean(\{r_{1},\cdots,r_{G}\})}}{{\rm std}(\{r_{1},\cdots,r_{G}\})}.
   \label{equation:grpo_adv}
\end{equation}

To prevent policy drift, a KL-divergence regularization term constrains the updated policy to remain close to the initial policy from Stage 1:
\begin{equation}
    D_{KL}(\pi_{\theta}|| \pi_{ref})=\frac{\pi_{ref}(o_{i}|x)}{\pi_{\theta}(o_{i}|x)}-\log\frac{\pi_{ref}(o_{i}|x)}{\pi_{\theta}(o_{i}|x)}-1.
\end{equation}

After reinforcement learning, we encode the entire candidate pool $\Corpus$ using the trained candidate encoder $\Encoder_{c}$. For a test input $x_{\rm test}$, retrieval follows a greedy decoding strategy, where at each step $t$, we select the candidate with the highest probability under the learned policy $\pi_{\theta}(\cdot|x_{\rm test}, z_{i<t})$. We find that beam search provides only marginal improvements for sequential retrieval, a similar observation noted in \citet{liu2024se2sequentialexampleselection}.
The retrieval process continues until $k$ examples are collected, forming the context for LLM inference. Evaluation is then performed on the LLM’s final generated program.

\section{Experiments}
\label{sec:exp}

\begin{table*}[!ht]
    \centering
    \small
    \setlength{\abovecaptionskip}{0.2cm}
    \setlength{\belowcaptionskip}{-0.4cm}
    \addtolength{\tabcolsep}{-3.5pt}
    \begin{threeparttable}
\begin{tabular}{@{}l
                S[table-format=2.2] 
                S[table-format=2.2]
                S[table-format=2.2]
                S[table-format=2.2]
                S[table-format=2.2]
                S[table-format=2.2]
                S[table-format=2.2]
                S[table-format=2.2]
                S[table-format=2.2]@{}}
\toprule
 & \multicolumn{8}{c}{\textbf{GeoQuery}} & \textbf{COVR-10} \\
 \cmidrule[0.4pt](r{0.125em}){2-9}%
 \cmidrule[0.4pt](lr{0.125em}){10-10}%
 & {i.i.d.} & {Template 1} & {Template 2} & {Template 3} & {TMCD 1} & {TMCD 2} & {TMCD 3} & {Length} & {avg.} \\ 
\midrule
Cover-LS \citep{levy-etal-2023-diverse} & 70.97 & 64.05 & 46.67 & 59.88 & 59.57 & 52.89 & 65.05 & 49.24 & 66.72 \\ 
\midrule
\multicolumn{10}{c}{\textit{Top-$k$ Retrieval}}\\ \midrule
Random       & 11.43 &  9.34 &  8.46 &  7.88 &  9.09 &  8.48 &  6.67 &  3.03 & 24.56 \\
BM25         & 72.14 & 40.36 & 26.89 & 44.85 & 46.67 & 42.42 & 48.79 & 27.88 &  1.74 \\
BERT         & 77.50 & 43.97 & 26.28 & 45.75 & \textbf{55.45} & 41.81 & 54.24 & 29.09 & 25.14 \\
Contriever   & 69.28 & 41.86 & 27.79 & 30.30 & 50.30 & 40.61 & 53.33 & 30.91 & 28.17 \\
EPR \citep{rubin2022learning} & 67.50 & 44.28 & 26.89 & 35.76 & 45.45 & 40.91 & 53.03 & 27.88 & 27.82 \\ 
\midrule
\multicolumn{10}{c}{\textit{Sequential Retrieval}}\\ \midrule
$se^{2}$ \citep{liu2024se2sequentialexampleselection}  & 74.28 & 30.42 & 26.28 & 41.21 & 49.69 & 42.12 & 54.54 & 31.51 & 29.46 \\
$se^{2} +$ tri-encoder & 77.50 & 54.81 & 26.58 & 43.03 & 50.61 & 43.03 & 55.15 & 31.52 & 30.44 \\
\method w/o SFT (Ours)  & 72.86 & 39.76 & 25.68 & 30.91 & 46.97 & 39.39 & 53.64 & 28.79 & 26.72 \\
\method w/o RL (Ours)   & 77.85 & 58.73 & 28.39 & 45.76 & 50.61 & 43.94 & 57.58 & 32.12 & 33.64 \\
\method (Ours)         & \textbf{78.21} & \textbf{59.64} & \textbf{33.83} & \textbf{48.48} & 51.52 & \textbf{44.24} & \textbf{59.09} & \textbf{33.03} & \textbf{36.18} \\
\bottomrule
\end{tabular}
    \end{threeparttable}
\caption{Exact-match accuracy on GeoQuery and COVR-10. Results highlight the advantages of a tri-encoder design and compositional retrieval via a Markov Decision Process.}
\label{tab:main}
\end{table*}

\subsection{Datasets}
We conduct our experiments on several Compositional Generalization tasks for generating semantic parsing program. This task inherently involves generating new combinations of familiar structures and symbols, explicitly highlighting the compositional nature. This explicitness makes it an ideal initial scenario to examine how context compositions impact model generation capabilities.

\paragraph{GeoQuery.} GeoQuery \citep{zelle1996learning,tang2001using} is a corpus of 880 questions related to U.S.\ geography. We follow the splits introduced by \citet{shaw-etal-2021-compositional}, which include: 1) Template split: Target programs are anonymized into templates, then randomly partitioned into training and test sets \citep{finegan-dollak-etal-2018-improving}.
2) TMCD split: The distribution of compounds in the training data is made maximally different from that of the test data \citep{keysers2020measuring}.
3) Length split: The test set contains longer program sequences than the training set.
\paragraph{COVR-10.} COVR-10 \cite{bogin-etal-2022-unobserved} is a synthetic dataset featuring a variable-free functional language. It comprises 10 compositional grammar splits, each of which includes distinct local structures in the test set that do not appear in training. Following prior setups \citep{levy-etal-2023-diverse}, we aggregate results by averaging across the 10 splits.
More details are provided in Appendix \ref{sec:appendixA}.

\subsection{Experiment Setup}


\paragraph{Models} We initialize all three encoders in our proposed retriever with \verb|bert-base-uncased| embeddings \cite{devlin-etal-2019-bert}. To compare with previous results, however, Codex (\verb|code-davinci-002|) \citep{chen2021evaluating,ouyang2022training} has been deprecated. As recommended by OpenAI, we replace it with \verb|gpt-3.5-turbo-instruct| for our experiments.

\paragraph{Evaluation} Following prior work, we use exact match accuracy as the main metric for evaluation. Results are averaged over 3 random seeds unless stated otherwise.

\paragraph{Model setup}
In our experiments, we set the total number of steps to $k=4$, which represents the number of examples within the context. For coverage calculation, we consider a maximum local structure size of $l=4$. In the proposed tri-encoder used in RCR, we set $\lambda=0.1$ in Eq.~\eqref{equation:logits}. The scaling temperature $\tau$ for computing the RL policy is set to $0.2$. The clip ratio $\epsilon$ in GRPO is set to $0.2$, and the coefficient $\beta$ of the KL divergence is set to $0.04$ (Eq.~\eqref{equation:grpo}).

\paragraph{Training details}
In the SFT stage, we construct the hard negatives set by retrieving the top-50 examples using BM25 and selecting the five examples with the lowest coverage for the instance $(x,\Candidates_{t-1})$. From these five examples, one hard negative is randomly sampled. We use a batch size of 64 for SFT and train for 120 epochs with a learning rate of $1\times10^{-5}$, employing a linear learning rate decay scheduler. 
In the RL stage, we train the model with a batch size of 16 and a group size of 32 in GRPO. The model is trained for 8 epochs with a learning rate of $1\times10^{-6}$.



\subsection{Baselines} 
We evaluate our method against a variety of baselines using the same inference LLM but different retrieval strategies. Both learning-free and learning-based retrievers are considered, including top-$k$ and sequential approaches.

\paragraph{Cover-LS.} Cover-LS \citep{levy-etal-2023-diverse} is an oracle setting in which the gold program is assumed to be accessible at test time. A BM25 retriever \citep{robertson2009probabilistic} then scores and selects examples covering the symbols of the gold program.

\paragraph{Top-$k$ methods.} 
\textbf{Random} imply samples $k$ examples from the candidate pool $\Corpus$. 
\textbf{BM25} \citep{robertson2009probabilistic} is a sparse retriever that scores candidate matches based on lexical overlaps with the query.
\textbf{BERT} uses a \verb|bert-base-uncased| encoder \citep{devlin-etal-2019-bert} to embed both the query and candidate examples, retrieving the top-$k$ by similarity (Eq.\ \ref{equation:dense_single}).
\textbf{Contriever} \citep{izacard2022unsupervised} follows a similar setup to BERT but replaces the encoder with a pre-trained “facebook/contriever” model that is trained in an unsupervised manner.
\textbf{EPR} \citep{rubin2022learning} is a learned retriever specifically for in-context example selection, where positive and negative pairs are sampled from a ranked candidate set (scored by a scoring LM’s log-likelihood of the ground-truth output).

\paragraph{Sequential retrieval methods} $\mathbf{\textit{se}^{2}}$ \citep{liu2024se2sequentialexampleselection} is a bi-encoder retriever that sequentially selects in-context examples, leveraging a large scoring LM to rank candidates. Unlike EPR, it constructs training data by iteratively extending the context sequence. We also adapt this method to our tri-encoder architecture (denoted as ``$se^{2}$+tri-encoder''), while keeping all other $se^{2}$ settings unchanged. 
\textbf{Reinforcing Compositional Retrieval (RCR)} is our proposed method, trained via two stages (SFT and RL). We also report ablations: ``RCR w/o SFT'' and ``RCR w/o RL''.

\paragraph{RL variations} We compare different advantage estimations $A_{i}$ in Eq.~\eqref{equation:grpo}. \textbf{NB} denotes reinforce without baseline, directly uses the reward $r(i)$ for training ($A_{i}=r_{i}$). \textbf{Remax} \citep{li2024remax} estimates a baseline by greedily sampling a sequence $o^{\prime}\in\argmax\pi_{\theta}(\cdot|x)$ and calculating the associated reward value, the advantage is $A_{i}=r_{i}-r(o^{\prime})$. \textbf{RLOO} \citep{ahmadian2024rloo} computes an unbiased return estimate by removing the contribution of each sampled sequence from the group, the advantage is calculated by $A_{i}=r_{i}-\frac{1}{k-1}\sum_{j\neq i}r_{j}$. \textbf{GRPO} applies the group advantage formulation from Eq.~\eqref{equation:grpo_adv}. While GRPO directly includes a KL-divergence term in its objective (Eq.\ \ref{fig:grpo}), we implement the other variants adding KL penalty to the reward \citep{ouyang2022training}.

\subsection{Results and Analysis}
\begin{table}[t]
    \footnotesize
    \addtolength{\tabcolsep}{-0.7mm}
    \renewcommand\arraystretch{1.2}

        \centering
        \begin{tabular}{lccccc}
        \toprule
            Dataset & w/o RL & NB & Remax & RLOO & GRPO\\
            \midrule
            i.i.d.  & 77.85 & 68.57 & 72.14 & 75.71 & \textbf{78.21}\\
            Template 1 & 58.73 & 10.84 & 30.12 & 55.72 & \textbf{59.64}\\
            Template 2 & 28.39 & 6.04 & 19.63 & 26.58 & \textbf{33.83}\\
            Template 3 & 45.76 & 28.48 & 29.39 & 48.18 & \textbf{48.48}\\
            TMCD 1 & 50.61 & 44.54 & 47.57 & 50.30 & \textbf{51.52}\\
            TMCD 2 & 43.94 & 37.87 & 43.63 & 42.12 & \textbf{44.24}\\
            TMCD 3 & 57.58 & 51.51 & 55.75 & 58.18 & \textbf{59.09}\\
            Length & 32.12 & 29.09 & 28.48 & 28.78 & \textbf{33.03}\\
        \bottomrule
        \end{tabular}
    \caption{Comparison of various RL advantage estimators on the GeoQuery dataset. GRPO consistently achieves the highest accuracy across all splits.}
    \label{tab:rl}
\end{table}

\paragraph{Main results and ablation studies.}
Table \ref{tab:main} presents our main experimental results. Overall, our proposed RCR method consistently outperforms both top-$k$ and sequential baselines on most splits, underscoring the importance of explicitly modeling inter-example dependencies through a sequential selection framework. We further analyze the contribution of our two-stage training pipeline via two ablations:
1) ``RCR w/o RL'' omits the RL phase, relying solely on SFT. Although this setting already surpasses many baselines, adding RL yields a significant performance boost, highlighting the impact of task-specific rewards in refining the policy and aligning the retriever with the LLM’s preferences.
2) ``RCR w/o SFT''. Here, we eliminate SFT and directly train with RL. Although competitive with certain baselines, its weaker results suggest that sequential retrieval benefits substantially from a well-initialized policy.

\begin{figure}[t]
    \centering
    \includegraphics[width=0.8\linewidth]{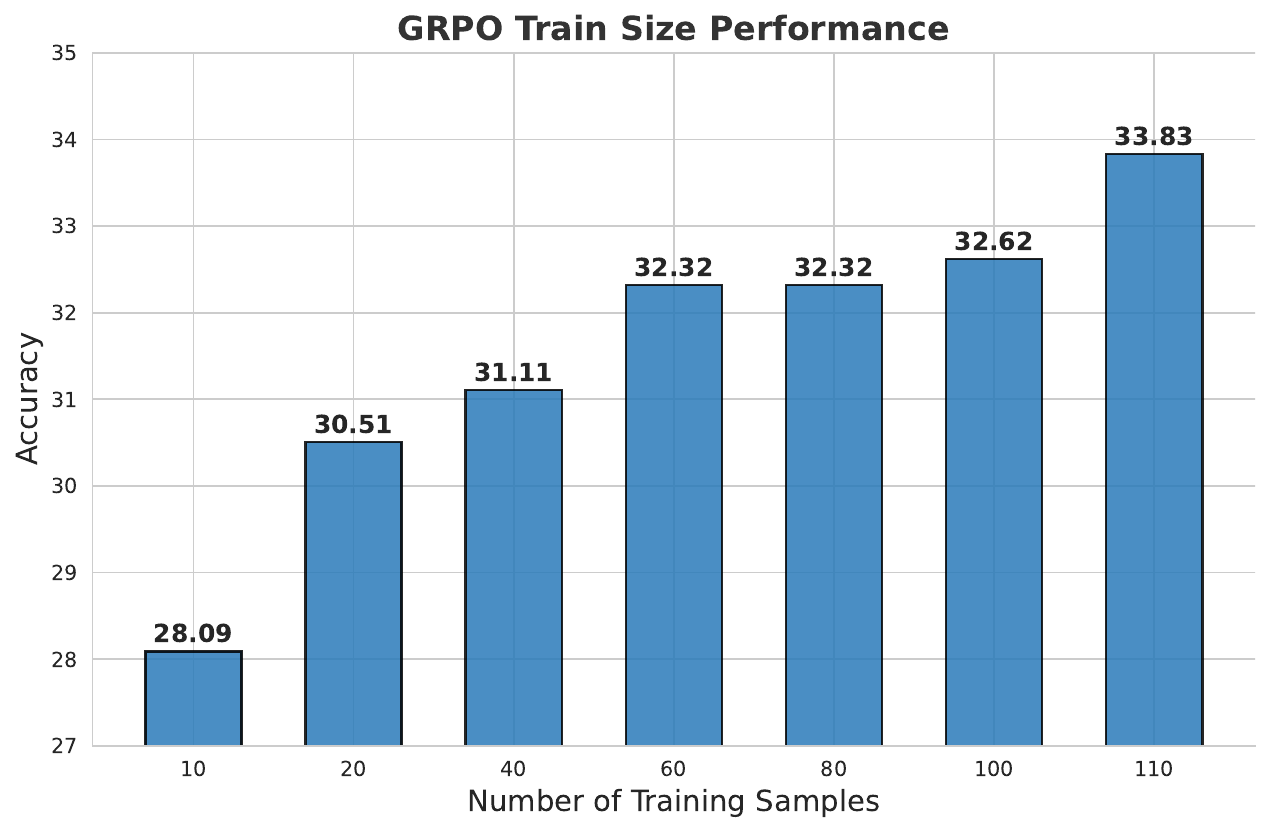}
    \caption{Performance over various numbers of training samples in GRPO. Results are evaluated in GeoQuery-Template2.}
    \label{fig:train_size}
\end{figure}

\paragraph{Tri-encoder architecture brings notable gains.}
We also compare ``$se^{2}$+ tri-encoder'' with $se^{2}$ to assess the impact of our tri-encoder architecture. Adopting the tri-encoder in  $se^{2}$ markedly improves results, suggesting that separating the encoding of query and retrieved examples is beneficial, especially for program parsing, where appending retrieved programs directly to the query can obscure the original query signal and promote repetitive selection (e.g., selecting items that mirror already retrieved examples due to the addition to the context of original query).
Moreover, while $se^{2}$ relies on GPT-Neo-2.7B \citep{black2022gptneo} to rank candidates, our approach leverages an efficient data construction strategy based on maximizing sub-structure coverage, forgoing the computational overhead of scoring-based systems. As evidenced by the comparison of ``RCR w/o RL'' versus ``$se^{2}$+ tri-encoder'', our method achieves higher accuracy under a more efficient training paradigm.

\paragraph{Normalized advantage yields greater stability.}
Table \ref{tab:rl} compares different advantage-estimation strategies for reinforcement learning. We observe that GRPO consistently provides the highest accuracy across all GeoQuery splits, indicating that its group-based advantage formulation is more stable and robust compared to alternative methods. By contrast, NB uses raw rewards without a baseline, yields noticeably lower scores, particularly on the more challenging Template 1 and 2 splits.

\paragraph{Impact of training sample size in RL stage}
Figure~\ref{fig:train_size} shows how accuracy varies with the number of training samples in our RL stage.
Initially, a significant improvement is observed as the number of training samples increases from 10 to 40, demonstrating the strong impact of additional data in the early stages. 
performance gains become more incremental, with accuracy stabilizing between 60 and 100 samples, where additional data contributes less noticeably.
Further increasing the training size to 100 and 110 samples still leads to continued progress. This pattern indicates that the RL stage will continue to benefit from additional training data, albeit at a diminishing rate.

\paragraph{Impact of group size in GRPO}
\begin{figure}[t]
    \centering
    \setlength{\abovecaptionskip}{0.2cm}
    \setlength{\belowcaptionskip}{-0.4cm}
    \includegraphics[width=0.8\linewidth]{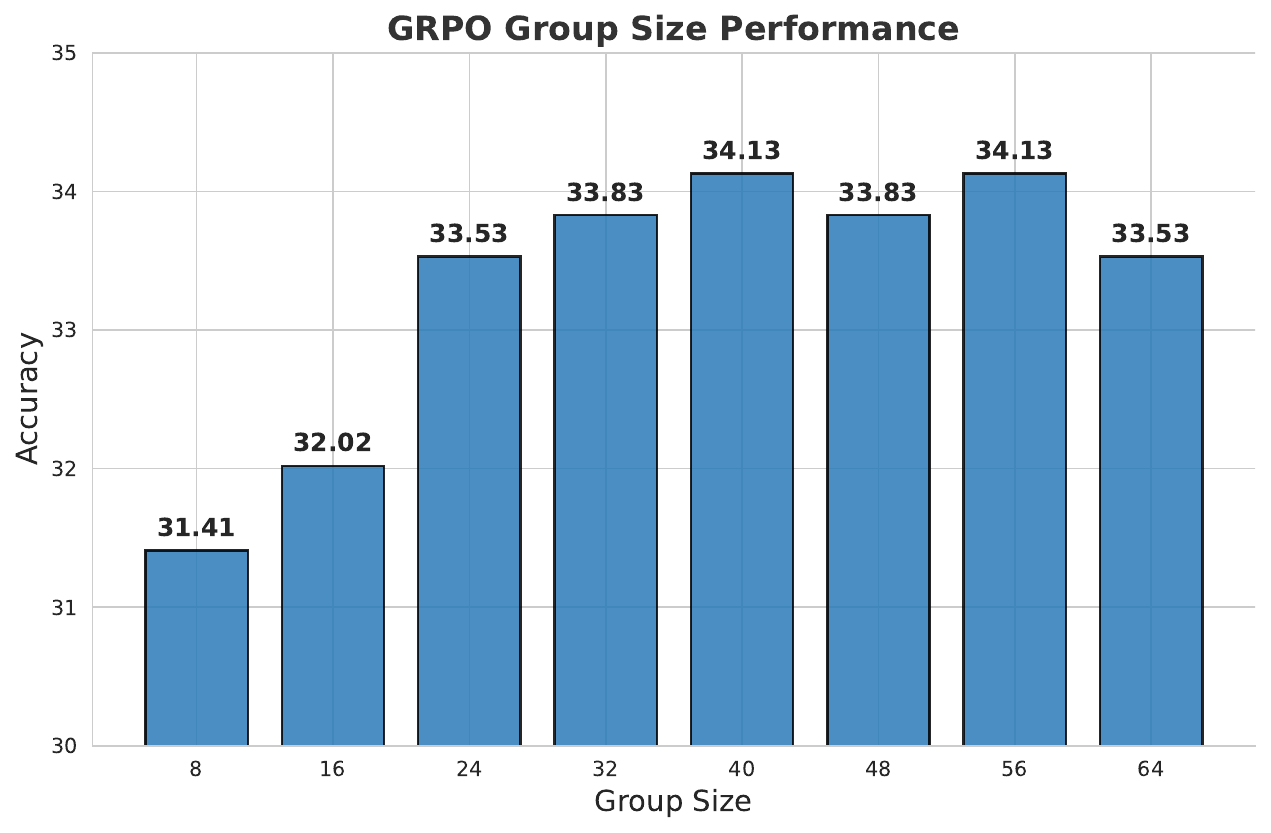}
    \caption{Performance over various group size in GRPO. Results are evaluated in GeoQuery-Template2.}
    \label{fig:group_size}
\end{figure}

The experiment results in Figure~\ref{fig:group_size} show that in GRPO, accuracy improves with larger group sizes up to a threshold, after which gains diminish. Smaller groups perform worse, while moderate sizes yield significant improvements. Beyond an optimal size, further increases offer little benefit and may introduce inefficiencies. Balancing group size for both accuracy and computational efficiency is key to maximizing GRPO performance in our experiments.


\section{Related Work}
\label{sec:related}
Early example selection methods retrieve semantically similar demonstrations \citep{liu2022makes}, but semantic proximity alone does not always yield optimal exemplars.
To improve selection, \citet{rubin2022learning} propose training a dense retriever with a scoring LM, ranking candidates by the log-likelihood of the ground-truth output. Extensions further refine this retriever across multiple tasks \citep{li2023unified,wang-etal-2024-learning}. Other studies highlight the importance of diversity in retrieval, such as structure or label diversity \citep{an2023context,long2024does,levy-etal-2023-diverse,pmlr-v202-ye23c}. However, many of these approaches treat each candidate independently, overlooking inter-example dependencies that influence ICL effectiveness.

Recent empirical studies~\citep{an2023context,long2024does} reveal that \textbf{diversity} (e.g., structure diversity, label diversity) also play an important role in demonstration selection. \citet{levy-etal-2023-diverse} select diverse demonstrations that aims to collectively cover all of the required structures in the semantic parsing program. \citet{pmlr-v202-ye23c} learns to select diverse example set based on the conditional determinantal point process.
However, previous works~\citep{rubin2022learning,wang-etal-2024-learning} overlook the interplay between in-context examples as they treat each candidate example independently. This may be suboptimal as the in-context examples can influence each other, previous work indicate that ICL is sensitive to the order of in-context examples~\citep{lu-etal-2022-fantastically}. To capture the \textbf{exemplar dependency}, recent works~\citep{liu2024demorank} and~\citep{ liu2024se2sequentialexampleselection} propose a data construction method by sequentially select and extend the in-context sequence from a LLM, the constructed dependency-aware data will be leverage to train a retriever via contrastive loss.

Different from indirect method that learning dependency via constructed training data, \citet{zhang-etal-2022-active} directly frame ICL problem as a decision making problem and use offline Q-learning to actively select examples from unlabeled set to label. 
However, this method does not learn a retriever to acquire effective examples for each test instance, during the inference, this method will traverse the entire training set to greedly choose an action based on the Q-value which is produced by a scoring LM as well. Therefore, this method is costly when action space (i.e., training dataset) is large and it lacks diversity. Although our method is also decision making, our method presents key differences with \citet{zhang-etal-2022-active}, as our method is retrieval-based, allowing sample and inference efficiency and to explore diversity. A similar work framing retrieval as MDP is proposed by \citet{chen-etal-2024-learning-retrieve}, their reward is primarily likelihood-oriented (white-box), whereas ours focuses on response-oriented (black-box). Besides, our reward design focus heavily on structural features which is crucial for program generation.

A comprehensive comparison of related methods and their properties is provided in Appendix \ref{sec:appendixC}, with a detailed summary in Table \ref{tab:summary}.

\section{Conclusion}
\label{sec:conclusion}
In this work, we introduce compositional retrieval, a retrieval paradigm that models the selection of multiple interdependent examples. To achieve this, we propose a tri-encoder sequential retriever, formulating retrieval as a MDP and training it in two stages SFT via efficient data construction and RL for policy refinement. Experiments highlight the potential of compositional retrieval for tasks requiring multiple sources of evidence and offer insights into the benefits and limitations of RL in retrieval.

\section{Limitations}
\label{sec:limitation}
While our approach effectively models compositional retrieval, it has several limitations. First, similar to generative retrieval, our retriever cannot dynamically incorporate new entries after training. Second, as retrieval steps increase, computational costs grow, and learning stability decreases due to compounding errors in sequential selection. To mitigate this, we focus on retrieval with a small number of steps, balancing efficiency and effectiveness. Future work could explore scalable retrieval mechanisms that maintain stability over longer sequences.



\bibliography{custom}

\appendix
\onecolumn

\section{Dataset details}
\label{sec:appendixA}
\begin{table}[ht]
\centering
\normalsize
\addtolength{\tabcolsep}{1pt}
\begin{tabular}{@{}llccc@{}}
\toprule
Dataset & Split & \multicolumn{1}{c}{Train} & \multicolumn{1}{c}{Development} & \multicolumn{1}{c}{Test} \\ \midrule
\multirow{8}{*}{GeoQuery} & Standard & 600 & - & 280 \\
 & Template1 & 438 & 110 & 332 \\
 & Template2 & 439 & 110 & 331 \\
 & Template3 & 440 & 110 & 330 \\
 & TMCD1 & 440 & 110 & 330 \\
 & TMCD2 & 440 & 110 & 330 \\
 & TMCD3 & 440 & 110 & 330 \\
 & Length & 440 & 110 & 330 \\ \midrule
COVR-10 & Each split & 3000 & - & 500 \\ \bottomrule
\end{tabular}%
\caption{Dataset sizes}
\label{tab:dataset_sizes}
\end{table}

\paragraph{GeoQuery.} GeoQuery \citep{zelle1996learning,tang2001using} is a widely used semantic parsing dataset containing 880 natural language questions about U.S.\ geography. Each query is labeled with a corresponding logical form that captures the precise meaning of the question. Following \citet{shaw-etal-2021-compositional}, we adopt three different data splits designed to test various aspects of compositional generalization:
\begin{itemize}
    \item Template Split: In this split, target logical forms are anonymized into templates, and then these templates are randomly partitioned between the training and test sets \citep{finegan-dollak-etal-2018-improving}. This procedure ensures that the model cannot simply memorize specific entity names and must instead learn the underlying template patterns.
    \item TMCD Split: The TMCD split \citep{keysers2020measuring} is constructed so that the distribution of compounds (sub-expressions in the logical forms) in the training set differs significantly from that in the test set. This presents a more challenging scenario, as the model must generalize to unseen or underrepresented logical substructures that do not appear (or rarely appear) in the training examples.
    \item Length Split: Here, test set queries and their corresponding logical forms are systematically longer than those in the training set. By ensuring a length discrepancy, this split evaluates whether the model can generalize to deeper or more complex logical forms that exceed the length of any training example.
\end{itemize}
While some previous works average performance results across multiple runs or across different TMCD/template variants, we present individual split results to show precisely how our model behaves in each compositional scenario.

\paragraph{COVR-10.}
COVR \citep{bogin-etal-2022-unobserved} is a synthetic dataset designed for testing compositional generalization in a variable-free functional language. Unlike GeoQuery, which focuses on a real-world domain (U.S.\ geography) with a relatively small set of queries, COVR generates a more diverse and systematically controlled set of expressions.

In particular, COVR-10 consists of 10 different compositional grammar splits, each crafted such that the test set includes certain local structures (sub-functional forms) that never appear during training. This forces a model to extrapolate beyond the specific structures seen in training and handle previously unseen or novel combinations of language constructs. Across these splits, the dataset’s design systematically varies the presence or arrangement of operations like \emph{largest}, \emph{filter}, or \emph{logical connectives} in the target expressions. Each split highlights a different angle of compositional difficulty. By evaluating on all 10 splits individually, we can observe how robustly our approach adapts to a range of compositional challenges.

Statistics are provided in Table. \ref{tab:dataset_sizes}.

We use 110 samples for the GRPO training stage across all benchmarks for a fair comparison. Specifically, we split the training dataset into two subsets: 110 samples for the GRPO stage and the remaining samples as retrieval candidate pool. Due to the limited number of total samples in the GeoQuery dataset (e.g., only 440 samples in GeoQuery Template 3, TMCD, and Length tasks as indicated in Table. \ref{tab:dataset_sizes}), increasing the GRPO training sample size would significantly reduce the candidate set, adversely impacting retrieval performance. Therefore, we maintained 110 GRPO training samples to ensure sufficient candidates remain available for effective retrieval.



\section{Local Structure}
\label{sec:appendixB}

\begin{table*}[!ht]
\centering
\footnotesize
\begin{tabular}{p{2.95cm}p{12.1cm}}
\toprule[2pt]
\textbf{Dataset} & COVR-10 \\
\textbf{Utterance} & \textit{What is the number of black mouse that is playing with dog ?} \\
\textbf{Program} & \texttt{count ( with\_relation ( filter ( black , find ( mouse ) ) , playing with , find ( dog ) ) )} \\

\textbf{Anonymized Program} & \texttt{count ( with\_relation ( filter ( ANON\_TYPE\_VALUE , find ( ANON\_ENTITY ) ) , ANON\_RELATION , find ( ANON\_ENTITY ) ) )} \\
\toprule
{\bf Size} & {\bf Local structures} \\
 \midrule
 \multirow{9}{*}{1} & 
 \texttt{count } \\ 
 & \texttt{with\_relation } \\
 & \texttt{filter} \\ 
 & \texttt{black} \\  
 & \texttt{find} \\ 
 & \texttt{mouse } \\
 & \texttt{playing } \\
 & \texttt{with } \\
 & \texttt{dog } \\
 \midrule
 \multirow{13}{*}{2} & 
 \texttt{<root> $\rightarrow$ count} \\
& \texttt{count $\rightarrow$ with\_relation} \\
& \texttt{filter $\rightarrow$ black} \\
& \texttt{filter $\rightarrow$ find} \\
& \texttt{find $\rightarrow$ dog} \\
& \texttt{find $\rightarrow$ mouse} \\
& \texttt{playing $\rightarrow$ with} \\
& \texttt{with\_relation $\rightarrow$ filter} \\
& \texttt{with\_relation $\rightarrow$ find} \\
& \texttt{with\_relation $\rightarrow$ playing} \\
& \texttt{black $\leftrightarrow$ find} \\
& \texttt{filter $\leftrightarrow$ playing} \\
& \texttt{playing $\leftrightarrow$ find} \\
\midrule
\multirow{13}{*}{3} & 
\texttt{<root> $\rightarrow$ count $\rightarrow$ with\_relation} \\
& \texttt{count $\rightarrow$ with\_relation $\rightarrow$ filter} \\
& \texttt{count $\rightarrow$ with\_relation $\rightarrow$ find} \\
& \texttt{count $\rightarrow$ with\_relation $\rightarrow$ playing} \\
& \texttt{filter $\rightarrow$ find $\rightarrow$ mouse} \\
& \texttt{filter $\rightarrow$ black $\leftrightarrow$ find} \\
& \texttt{with\_relation $\rightarrow$ filter $\rightarrow$ black} \\
& \texttt{with\_relation $\rightarrow$ filter $\rightarrow$ find} \\
& \texttt{with\_relation $\rightarrow$ find $\rightarrow$ dog} \\
& \texttt{with\_relation $\rightarrow$ playing $\rightarrow$ with} \\
& \texttt{with\_relation $\leftrightarrow$ filter $\leftrightarrow$ playing} \\
& \texttt{with\_relation $\rightarrow$ playing $\leftrightarrow$ find} \\
& \texttt{filter $\leftrightarrow$ playing $\leftrightarrow$ find} \\
  \midrule
  \multirow{12}{*}{4} & 
\texttt{<root> $\rightarrow$ count $\rightarrow$ with\_relation $\rightarrow$ filter} \\
& \texttt{<root> $\rightarrow$ count $\rightarrow$ with\_relation $\rightarrow$ find} \\
& \texttt{<root> $\rightarrow$ count $\rightarrow$ with\_relation $\rightarrow$ playing} \\
& \texttt{count $\rightarrow$ with\_relation $\rightarrow$ filter $\rightarrow$ black} \\
& \texttt{count $\rightarrow$ with\_relation $\rightarrow$ filter $\rightarrow$ find} \\
& \texttt{count $\rightarrow$ with\_relation $\rightarrow$ find $\rightarrow$ dog} \\
& \texttt{count $\rightarrow$ with\_relation $\rightarrow$ playing $\rightarrow$ with} \\
& \texttt{count $\rightarrow$ with\_relation $\rightarrow$ filter $\leftrightarrow$ playing} \\
& \texttt{count $\rightarrow$ with\_relation $\rightarrow$ playing $\leftrightarrow$ find} \\
& \texttt{with\_relation $\rightarrow$ filter $\rightarrow$ find $\rightarrow$ mouse} \\
& \texttt{with\_relation $\rightarrow$ filter $\rightarrow$ black $\leftrightarrow$ find} \\
& \texttt{with\_relation $\rightarrow$ filter $\leftrightarrow$ playing $\leftrightarrow$ find} \\
 \bottomrule[2pt]
\end{tabular}
\caption{Local structures of different sizes for a specific example ($\rightarrow$ denotes parent-child relations, $\leftrightarrow$ denotes sibling relations)}
\label{tab:ls_examples}
\end{table*}

Table~\ref{tab:ls_examples} demonstrates several sample local structures with size $\leq 4$ extracted from a program. We collect all such structures (from size 1 to 4) into a set, denoted by $LS^{4}(x)$.

Formally, following \citet{bogin-etal-2022-unobserved,levy-etal-2023-diverse}, we define \textit{local structures} as small, connected subgraphs of its parse tree. Given a target program $y$, we first convert it into a tree $T = (\mathcal{V}, \mathcal{E})$, where each node $v \in \mathcal{V}$ is labeled with a program symbol (e.g., a function, operator, or constant). The edge set $\mathcal{E}$ encodes parent-child relationships, corresponding to function--argument links in $y$.
Next, following \citet{levy-etal-2023-diverse}, we introduce sibling relations to form a graph $G = (\mathcal{V}, \mathcal{E} \cup \mathcal{E}_{\mathrm{sib}})$. Concretely, for each parent node $p$ in $T$ with children $(c^p_1,\dots,c^p_{N_p})$, we connect consecutive children with edges $(c^p_i, c^p_{i+1})$, thus collecting them in $\mathcal{E}_{\mathrm{sib}}$. This step reflects the intuition that sibling nodes often appear as parallel arguments for the same function or operator.

A local structure of size $n$ is then any connected subgraph $G_{\mathrm{LS}} \subseteq G$ with exactly $n$ nodes, satisfying the condition that any pair of nodes $(x,y)$ in $G_{\mathrm{LS}}$ is connected by a sibling edge in $\mathcal{E}_{\mathrm{sib}}$ if and only if both $x$ and $y$ are leaf nodes within $G_{\mathrm{LS}}$. In simpler terms, these subgraphs capture hierarchical (parent-child) and lateral (sibling) relationships while excluding more distant familial ties (e.g., ``cousins'' or ``uncles'' in the broader tree). 

\section{Comparison to Related Works}
\label{sec:appendixC}
\begin{table*}[t]
\centering
{\renewcommand{\arraystretch}{0.9}
\setlength{\tabcolsep}{2.5pt}
\small
\begin{tabular}{@{}l c c c c c c c@{}}
\toprule
{\bf Methods} & \makecell{{\bf Scoring}\\{\bf LM}} & \makecell{{\bf Black-box}\\{\bf LM}} & \makecell{{\bf Compositional}\\{\bf Optimization}} & \makecell{{\bf Reward}\\ {\bf Modeling}} & \makecell{{\bf Few-Shot}\\{\bf Training}} & \makecell{{\bf Exemplar}\\{\bf Dependency}} & {\bf Diversity} \\ \midrule
CSL-Aug~\citep{qiu-etal-2022-improving}  & \xmark & \xmark & \cmark & \xmark & \xmark & \xmark & \xmark\\
EPR~\citep{rubin2022learning} & \cmark & \xmark & \xmark & \xmark & \xmark & \xmark & \xmark \\
Active-RL~\citep{zhang-etal-2022-active} & \cmark & \xmark & \cmark & \xmark & \cmark & \cmark & \xmark \\
Cover-LS~\citep{levy-etal-2023-diverse} & \xmark & \xmark & \cmark & \xmark & \xmark & \xmark & \cmark \\
CEIL~\citep{pmlr-v202-ye23c} & \cmark & \xmark & \cmark & \xmark &  \xmark & \xmark & \cmark\\
LLM-R~\citep{wang-etal-2024-learning} & \cmark & \xmark & \xmark &  \cmark & \xmark & \xmark & \xmark\\ 
Se$^{2}$~\citep{liu2024se2sequentialexampleselection} & \cmark & \xmark & \xmark & \xmark & \xmark &  \cmark & \cmark\\
ITERR~\citep{chen-etal-2024-learning-retrieve} & \cmark & \xmark & \cmark & \cmark & \cmark & \cmark & \cmark \\

\midrule
\method ({\bf Ours}) & \cmark & \cmark & \cmark  & \cmark & \cmark &  \cmark & \cmark \\
\bottomrule
\end{tabular}
}
\caption{ 
Comparison of different in-context learning (ICL) methods in terms of several desirable properties. 
\textbf{Scoring-LM} means whether the in-context examples are ranked by a language model (LM) to obtain the training signal.
\textbf{Black-box LM} means whether the in-context examples are ranked by a language model (LM) without accessing the model probability.
\textbf{Compo Optim}, compositional optimizing method, which scores and optimizes over the entire in-context examples set and can generalize well according to compositionality.  
\textbf{Reward Modeling} trains a reward model to capture the differences of in-context examples and provide fine-grained LM preference.
\textbf{Few-Shot Training} means the example selection model is trained with only few data samples.
\textbf{Exemplar Dependency} means that selection of in-context examples is based on the conditional probability of already selected examples.
\textbf{Diversity} represents the method considers increasing the diversity of selected set of examples.
For detailed differences compared to previous works, see Section \ref{sec:related} and Appendix \ref{sec:appendixB} for more discussion.
}
\label{tab:summary}
\end{table*}

Table~\ref{tab:summary} summarizes key differences between our approach and several existing in-context learning (ICL) methods across multiple dimensions: whether the retriever is learned, whether a separate scoring LM is required, whether the method optimizes the entire composition of in-context examples, whether it uses reward modeling, whether it supports few-shot training, how it handles exemplar dependency, and whether it considers diversity in retrieval.

Early example selection strategies often rely on semantic similarity between the query and candidate examples \citep{liu2022makes}. However, semantic proximity alone may overlook other factors crucial for downstream performance (e.g., structural coverage in logic-based tasks). To address this, \citet{rubin2022learning} propose training a dense retriever guided by a \textbf{scoring LM}, where the log-likelihood of the ground-truth output ranks each candidate example.

Other research emphasizes \textbf{diversity} in retrieval.  \citet{levy-etal-2023-diverse} cover essential symbols in a semantic parsing program by selecting demonstration examples collectively, while \citet{pmlr-v202-ye23c} adopt a conditional determinantal point process to learn a diverse example set. Nonetheless, these methods primarily treat each candidate independently, so interactions among selected examples are often overlooked.

Recent work addresses this \textbf{exemplar dependency} by constructing sequential training data that captures how each newly selected example influences future choices. For instance, \citet{liu2024demorank} and \citet{liu2024se2sequentialexampleselection} iteratively extend in-context examples based on a large LM’s feedback, then train a retriever with contrastive loss on this data. Although this approach integrates exemplar interdependence into the training set, it still relies on a scoring LM at construction time and does not formulate retrieval as a direct decision process.

Our proposed method directly models compositional retrieval as a MDP, yielding a fully learned retriever that is both efficient and capable of inter-example conditioning. We do not require an external scoring LM, relying instead on a tri-encoder architecture and a reward signal based on local structure coverage. This design allows us to optimize over the entire in-context composition (instead of independently scoring each example) and supports training with few data samples. Furthermore, unlike prior work that treats example dependency indirectly through data construction alone, our method explicitly captures it in both the model architecture and the training objective. Finally, our approach naturally promotes diversity by sequentially selecting items to maximize coverage, thereby improving ICL effectiveness without incurring the computational overhead of large scoring LMs.

A similar work is proposed by \citet{chen-etal-2024-learning-retrieve},
while both works share the perspective of framing retrieval as MDP, our approaches differ in key aspects: First, \citet{chen-etal-2024-learning-retrieve} use a GRU-based retriever, while we adopt a tri-encoder architecture designed for better control of state and candidate representations during sequential retrieval.
Second, their reward is primarily likelihood-oriented (white-box), whereas ours focuses on task-specific structural correctness (black-box), our black-box setting is more general than white-box setting. Besides, our reward design focus heavily on structural features which is crucial for program generation.
Third, we adopt SFT and GRPO (a customized policy optimization), which complements their approach and reflects different priorities in training efficiency and stability.

\section{Example Cases of Tasks}
\label{sec:appendixD}

\begin{table}[h]
\centering
\small
\renewcommand\arraystretch{1.3}
\caption{Prompts examples for LLM inference.}
\label{tab:prompts}

\begin{tabular}{p{0.95\textwidth}}
\toprule[2pt]

\multicolumn{1}{c}{\textbf{Datasets}: Geoquery} \\ \midrule[0.2pt]
\textbf{Utterance}: \textit{what is the highest point in states bordering georgia}\\
\textbf{Gold Program}: \texttt{answer(highest(place(loc\_2(state(next\_to\_2(stateid('value')))))))}\\
\midrule[0.2pt]
Source: \textit{which capitals are in the states that border texas}\\
Target: \texttt{(capital(loc\_2(state(next\_to\_2(stateid('value'))))))}\\
Source: \textit{what is the largest city in states that border california} \\ 
Target: \texttt{answer(largest(city(loc\_2(state(next\_to\_2(stateid('value')))))))}\\
Source: \textit{what are the capitals of the states that border texas} \\
Target: \texttt{answer(capital(loc\_2(state(next\_to\_2(stateid('value'))))))} \\
Source: \textit{which state has the lowest point that borders idaho}\\
Target: \texttt{answer(state(loc\_1(lowest(place(loc\_2(next\_to\_2(stateid('value'))))))))} \\
Source: \textit{what is the highest point in states bordering georgia} \\
Target:

\\ \midrule[2pt]
\multicolumn{1}{c}{\textbf{Datasets}: COVR-10} \\ \midrule[0.2pt]
\textbf{Utterance}: \textit{Either there is round animal or the color of mouse that is looking at cat is equal to round}\\
\textbf{Gold Program}: \texttt{or ( exists ( filter ( round , find ( animal ) ) ) , eq ( query\_attr [ color ] ( with\_relation ( find ( mouse ) , looking at , find ( cat ) ) ) , round ) )}\\
\midrule[0.2pt]

Source: \textit{Either the shape of animal is equal to triangle or the shape of square cat that is looking at animal is equal to brown}\\
Target: \texttt{or ( eq ( query\_attr [ shape ] ( find ( animal ) ) , triangle ) , eq ( query\_attr [ shape ] ( with\_relation ( filter ( square , find ( cat ) ) , looking at , find ( animal ) ) ) , brown ) )}\\
Source: \textit{Either the color of square triangle animal that is looking at brown animal that is looking at gray mouse is equal to triangle or the shape of black gray cat is equal to square}\\
Target: \texttt{or ( eq ( query\_attr [ color ] ( with\_relation ( filter ( square , filter ( triangle , find ( animal ) ) ) , looking at , with\_relation ( filter ( brown , find ( animal ) ) , looking at , filter ( gray , find ( mouse ) ) ) ) ) , triangle ) , eq ( query\_attr [ shape ] ( filter ( black , filter ( gray , find ( cat ) ) ) ) , square ) )}\\
Source: \textit{Either the color of white mouse is equal to square or the shape of white dog that is looking at triangle dog is equal to the color of square triangle mouse}\\ 
Target: \texttt{or ( eq ( query\_attr [ color ] ( filter ( white , find ( mouse ) ) ) , square ) , eq ( query\_attr [ shape ] ( with\_relation ( filter ( white , find ( dog ) ) , looking at , filter ( triangle , find ( dog ) ) ) ) , query\_attr [ color ] ( filter ( square , filter ( triangle , find ( mouse ) ) ) ) ) )}\\
Source: \textit{Either the color of black cat is equal to gray or none of round black dog are looking at mouse that is looking at cat}\\
Target: \texttt{or ( eq ( query\_attr [ color ] ( filter ( black , find ( cat ) ) ) , gray ) , none ( filter ( round , filter ( black , find ( dog ) ) ) , with\_relation ( scene ( ) , looking at , with\_relation ( find ( mouse ) , looking at , find ( cat ) ) ) ) )}\\
Source: \textit{Either there is round animal or the color of mouse that is looking at cat is equal to round}\\
Target: \\
\bottomrule[2pt]
\end{tabular}
\end{table}

In Table \ref{tab:ls_examples}, we provide prompts example in few-shot ICL manner for each task, We add special prefixes ``source:'' and ``target:'' for retrieved source-target pairs and separate them with break lines.

\section{Case Study}
\label{sec:appendixE}
\begin{table*}[!h]
\centering
\small
\begin{tabular}{p{2.55cm}p{12.1cm}}
\toprule[2pt]
\textbf{Dataset} & GeoQuery \\
\textbf{Utterance} & \textit{what is the \colorbox{yellow}{highest point} \colorbox{green}{in states bordering georgia}} \\
\textbf{Gold Program} & \texttt{\colorbox{yellow}{answer(highest(place(}\colorbox{green}{loc\_2(state(next\_to\_2(stateid('value')))))))}} \\
\toprule
 \multirow{10}{*}{\bf BM25} &source: \textit{what is the highest point in the us}  \\ 
 &target: \colorbox{pink}{\texttt{answer(highest(place(loc\_2(countryid('value')))))}}\\
 &source: \textit{what is the highest point in the usa} \\
 &target: \colorbox{pink}{\texttt{answer(highest(place(loc\_2(countryid('value')))))}}\\
 &source: \textit{what states have no bordering state} \\
 &target: \texttt{answer(exclude(state(all), next\_to\_2(state(all))))}\\
 &source: \textit{what is the highest point in the united states} \\
 &target: \colorbox{pink}{\texttt{answer(highest(place(loc\_2(countryid('value')))))}}\\
 &source: \textit{what is the highest point in states bordering georgia} \\
 &target:\\

\toprule

\multirow{10}{*}{\bf EPR} &source: \textit{what is the highest point in montana}  \\ 
&target: \texttt{\colorbox{cyan}{answer(highest(place(loc\_2(stateid('value')))))}}\\
&source: \textit{what is the highest point in new mexico} \\
&target: \texttt{\colorbox{cyan}{answer(highest(place(loc\_2(stateid('value')))))}}\\
&source: \textit{what is the highest point in rhode island} \\
&target: \texttt{\colorbox{cyan}{answer(highest(place(loc\_2(stateid('value')))))}}\\
&source: \textit{what is the highest point in virginia} \\
&target: \texttt{\colorbox{cyan}{answer(highest(place(loc\_2(stateid('value')))))}}\\
&source: \textit{what is the highest point in states bordering georgia} \\
&target:\\

\toprule
 
\multirow{10}{*}{$\mathbf{\textit{se}^{2}}$} &source: \textit{what state that borders texas is the largest}  \\ 
&target: \texttt{answer(largest(\colorbox{lightgray}{state(next\_to\_2(stateid('value')))))}}\\
&source: \textit{what states border states that border mississippi} \\
&target: \texttt{answer(state(next\_to\_2(\colorbox{lightgray}{state(next\_to\_2(stateid('value')))))))}}\\
&source: \textit{what states border states that border states that border florida} \\
&target: \texttt{answer(state(next\_to\_2(state(next\_to\_2(\colorbox{lightgray}{state(next\_to\_2(stateid('value')))))))))}}\\
&source: \textit{what are the capitals of the states that border texas} \\
&target: \texttt{answer(capital(\colorbox{green}{loc\_2(state(next\_to\_2(stateid('value')))))))}}\\
&source: \textit{what is the highest point in states bordering georgia} \\
&target:\\

\toprule

\multirow{10}{*}{\bf RCR} &source: \textit{which capitals are in the states that border texas}  \\ 
&target: \texttt{answer(capital(\colorbox{green}{loc\_2(state(next\_to\_2(stateid('value'))))))}}\\
&source: \textit{what is the highest point in the us} \\
&target: \texttt{\colorbox{yellow}{answer(highest(place(}(loc\_2(countryid('value')))))}\\
&source: \textit{what is the largest city in states that border california} \\
&target: \texttt{answer(largest(city(\colorbox{green}{loc\_2(state(next\_to\_2(stateid('value')))))))}}\\
&source: \textit{what are the capitals of the states that border texas} \\
&target: \texttt{answer(capital(\colorbox{green}{loc\_2(state(next\_to\_2(stateid('value'))))))}}\\
&source: \textit{what is the highest point in states bordering georgia} \\
&target:\\

\bottomrule[2pt]
\end{tabular}
\caption{Program examples produced with various retrieval methods for a specific test example. Each prompt contains $k=4$ examples.}
\vspace{3cm}
\label{tab:prompt_examples}
\end{table*}

Table~\ref{tab:prompt_examples} presents various methods for retrieving program examples. BM25 and EPR are top-k retrieval methods designed to find the most similar examples. However, a key limitation of these methods is that they may retrieve examples containing the same program, leading to a lack of diversity and potentially failing to cover the gold program.
$\textit{se}^{2}$ employs sequential retrieval, selecting examples one by one, which introduces some level of diversity. Nevertheless, it still primarily focuses on retrieving similar examples and does not provide sufficient diversity to fully cover the gold program.
In contrast, our proposed method, RCR, leverages coverage-aware learning using SFT and RL to retrieve a more diverse set of examples, effectively improving coverage of the gold program.

\clearpage
\section{Additional Results}
\label{sec:appendixF}
\subsection{GRPO Training Data Size}
To analyze the effect of more GRPO training data, we ran experiments on the COVR-10 dataset. The results below show a trend where increasing the number of GRPO training samples improves performance initially and then gradually plateaus. This suggests that our method is capable of scaling with more training samples, but the marginal gains diminish after a certain point, indicating a convergence behavior. We believe this pattern aligns with typical reinforcement learning settings, where policy improvement benefits from more examples, but only up to the point where the additional data becomes less informative or redundant.

\begin{figure}[h]
    \centering
    \includegraphics[width=0.7\linewidth]{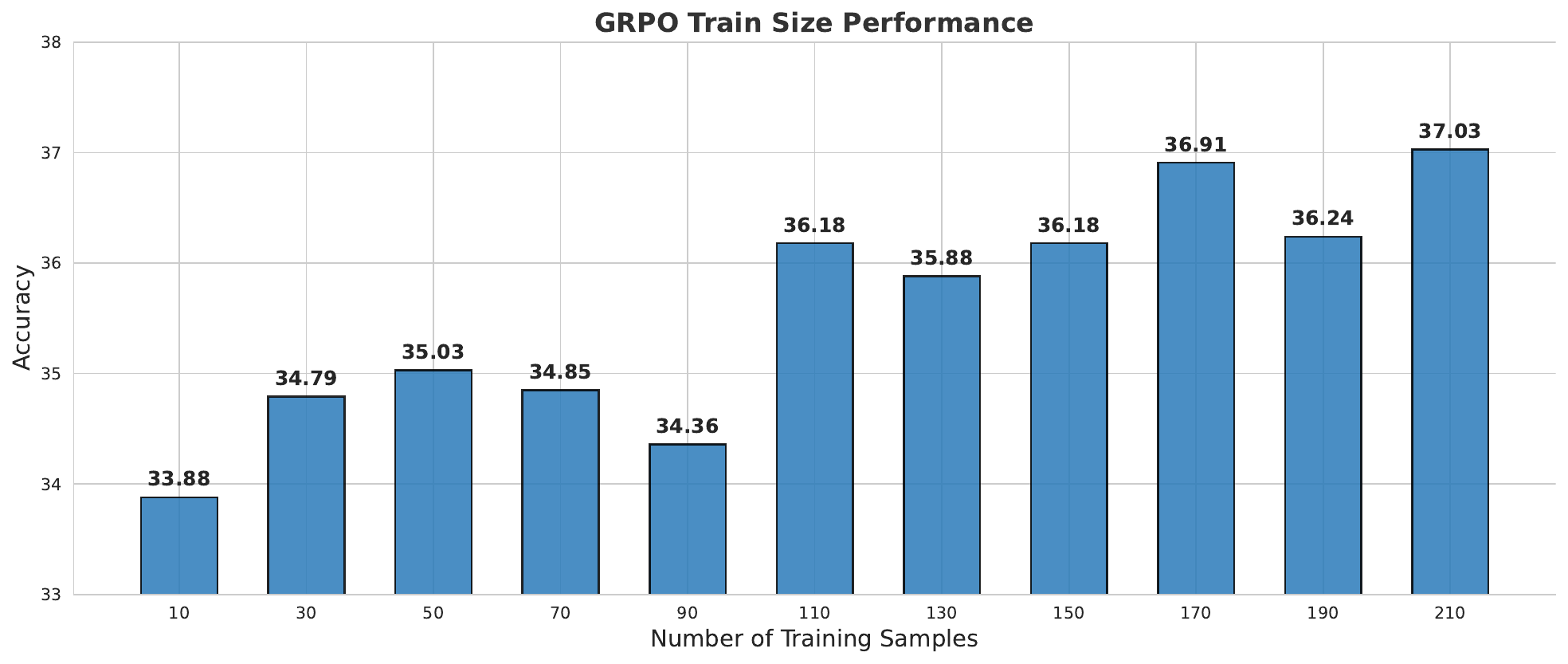}
    \caption{Performance over various numbers of training samples in GRPO. Results are evaluated in COVR-10.}
    \label{fig:train_size_covr}
\end{figure}

\subsection{Correctness-based Reward}
Program generation task enables the use of a continuous and informative reward (local structure coverage, Eq. (\ref{equation:reward})), significantly benefiting RL training and analysis. Although our main focus is Compositional Generalization tasks for generating semantic parsing program, we conduct additional experiments using a simpler, binary reward formulation $r\in\{0,1\}$, reflecting correctness of the model's final output. This discrete reward design aligns with use cases such as multi-hop question answering or classification tasks.

\begin{table*}[!h]
    \centering
    \small
    \addtolength{\tabcolsep}{-2.5pt}
    \begin{threeparttable}
\begin{tabular}{@{}rcccccccccc@{}}
\toprule
 & \multicolumn{8}{c}{\textbf{GeoQuery}} \\
 \cmidrule[0.4pt](r{0.125em}){2-9}%
 &  i.i.d. & Template 1 & Template 2 & Template 3 & TMCD 1 & TMCD 2 & TMCD 3 & Length \\ 
\midrule

Coverage reward & \textbf{78.21} & \textbf{59.64} & 33.83 & \textbf{48.48} & \textbf{51.52} & 44.24 & \textbf{59.09} & \textbf{33.03}  \\
Correctness reward & 77.85 & 59.03 & \textbf{36.85} & 47.87 & 50.60 & \textbf{45.15} & 58.48 & 32.42 \\

\bottomrule
\end{tabular}
\end{threeparttable}%

\caption{Exact-match accuracy on GeoQuery using different rewards.}
\label{tab:correctness_reward}
\end{table*}

Results in Table \ref{tab:correctness_reward} demonstrate our method’s robust performance, even with a binary correctness reward that is less informative than our original continuous reward, underscoring the broader applicability and generalization potential of our approach.

\subsection{Many-shot Setting for Compositional Generalization}
As larger context sizes enable LLMs to process extensive information directly, potentially reducing explicit retrieval needs.
To evaluate if the parsing program generation task can be solved by more in-context examples and longer contexts, we compare our retrieval-based approach against a many-shot setting in our ICL scenario. The results are summarized below:
\begin{table*}[h]
    \footnotesize
    \addtolength{\tabcolsep}{-0.7mm}
    \renewcommand\arraystretch{1.5}

        \centering
        \begin{tabular}{lccccc}
        \toprule
            \textbf{\#examples} & 10 & 20 & 100 & 150 & 4 (our trained RCR)\\
            \midrule
            i.i.d. split & 21.42 & 35.71 & 63.21 & 62.50 & \textbf{78.21} \\

        \bottomrule
        \end{tabular}
    \caption{Exact-match accuracy on GeoQuery i.i.d. split using different number of examples $k$ (affecting testing only).}
    \label{tab:many-shots}
\end{table*}

Table \ref{tab:many-shots} demonstrates that while increasing the number of in-context examples does improve performance up to a certain point (10 $\to$ 100 examples), further addition can negatively impact performance due to redundancy or repetitive information (as performance drops slightly from 63.21\% at 100 examples to 62.50\% at 150 examples). This observation aligns with our discussion in the main text where we note that repetitive or redundant examples may provide limited additional information or even degrade model performance in program generation.
Our retrieval-based method, by carefully selecting fewer but more informative examples (4 steps), significantly outperforms the many-shot long-context approach, highlighting the efficiency and effectiveness of explicit compositional retrieval.

\subsection{Trade-off Between Retrieval Accuracy and Efficiency}

We conducted a detailed analysis of how the number of retrieval steps impacts both performance and computational efficiency. Figure \ref{fig:steps} demonstrates our method’s performance across varying numbers of retrieval steps, as well as the time spent when evaluating the model on the entire test split:


            

\begin{figure}[h]
    \centering
    \includegraphics[width=0.55\linewidth]{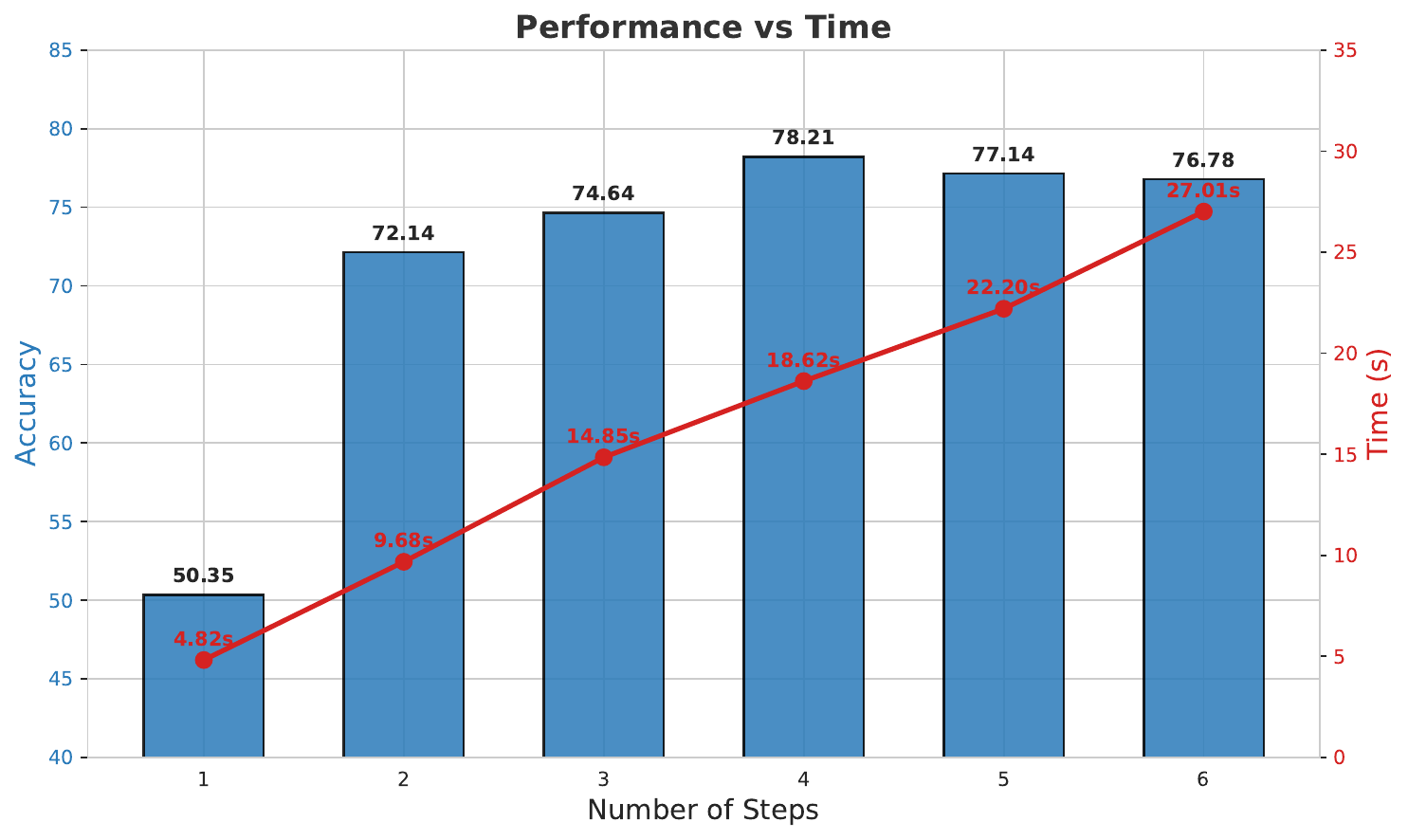}
    \caption{Exact-match accuracy on GeoQuery i.i.d. split using different number of steps $k$ (affecting both training and testing), we report the time spent when evaluating the model on the entire test split.}
    \label{fig:steps}
\end{figure}

As illustrated in Figure \ref{fig:steps}, our method achieves optimal performance with just 4 retrieval steps, which significantly reduces the computational overhead compared to longer sequences. Fewer steps directly translate to less computational time and lower costs.

Additionally, the specific design of our tri-encoder architecture further mitigates computational concerns. Notably, during sequential retrieval, as we seperately encode the input and retrieved candidates using two different encoder and add two embeddings at the top of the encoders, embeddings for the query need to be computed only once since the query remains unchanged throughout the retrieval process. Specifically:
\begin{itemize}
    \item At step 0, we compute the query embedding.
    \item At each subsequent step $t>0$, only embeddings for newly retrieved candidates at step $t-1$ are computed.
\end{itemize}
Therefore, the complexity of our retriever scales linearly as $O(n)$, where $n$ is the number of retrieval steps. This linear scaling ensures efficiency even when extending to more steps. In summary, our approach balances retrieval accuracy with computational efficiency by achieving optimal performance within very few steps. We appreciate the reviewer’s feedback, which allowed us to better highlight and clarify these important points.

\end{document}